\documentclass[sigconf]{acmart}
\AtBeginDocument{%
  \providecommand\BibTeX{{%
    \normalfont B\kern-0.5em{\scshape i\kern-0.25em b}\kern-0.8em\TeX}}}

\setcopyright{rightsretained} 
\copyrightyear{2021}
\acmYear{2021}
\acmDOI{} 

\acmConference[KDD WIT '21]{KDD '21: ACM SIGKDD International Conference on Knowledge Discovery \& Data Mining}{Aug 14--18, 2021}{Virtual Event, Singapore}
\acmBooktitle{Proceedings of ACM SIGKDD Workshop On Deriving Insights From User-Generated Text (KDD WIT'21), August 14–18, 2021, Virtual Event, Singapore}
\acmPrice{} 
\acmISBN{} 

\begin{document}

\title{Maps Search Misspelling Detection Leveraging Domain-Augmented Contextual Representations}


\author{Yutong Li}
\affiliation{%
  \institution{Apple Inc.}
  \city{Cupertino}
  \state{CA}
  \country{USA}
  \postcode{95014}}
\email{yutong_li@apple.com}

\renewcommand{\shortauthors}{Yutong Li}

\begin{abstract}
Building an independent misspelling detector and serve it before correction can bring multiple benefits to speller and other search components, which is particularly true for the most commonly deployed noisy-channel based speller systems. 
With rapid development of deep learning and substantial advancement in contextual representation learning such as BERTology\footnote{\url{https://huggingface.co/transformers/bertology.html}}, building a decent misspelling detector without having to rely on hand-crafted features associated with noisy-channel architecture becomes more-than-ever accessible. However BERTolgy models are trained with natural language corpus but Maps Search is highly domain specific, would BERTology continue its success. In this paper we design 4 stages of models for misspeling detection ranging from the most basic LSTM to single-domain augmented fine-tuned BERT. We found for Maps Search in our case, other advanced BERTology family model such as RoBERTa does not necessarily outperform BERT, and a classic cross-domain fine-tuned full BERT even underperforms a smaller single-domain fine-tuned BERT. We share more findings through comprehensive modeling experiments and analysis, we also briefly cover the data generation algorithm breakthrough. 
\end{abstract}




\maketitle

\section{Introduction}
Spell correction \cite{Kemighan90}\cite{Chiu14}\cite{Lai15}\cite{Jurafsky16}\cite{Jayanthi20}\cite{Hong19}\cite{Shaptala19}\cite{Zhang20} is a critical task for NLP and search application. The most common corrector is a ranker (such as XGBoost \cite{chen2016}) on top of noisy-channel architecture \cite{Kemighan90}\cite{Chiu14}\cite{Lai15}\cite{Jurafsky16}, leveraging language model \cite{Jurafsky16}\cite{Bergsma09}, error model \cite{Ahmad05}\cite{Brill2000} etc to rank a list of query candidates and select the top suggestion as correction. Language model and error model need to be pre-computed for populating hand-crafted features for the ranker/corrector, which usually involves heavy feature preparation and computation and various cache optimizations. What's more, since the corrector is not able to know beforehand whether a query is misspelt or not, it accepts all queries for processing. Considering misspelling ratio given a locale is usually between 10-20\% per our production analysis, which also aligns with the reported estimation 12\% \cite{Dalianis02}\cite{Martins04} for search engines, it means 80-90\% of the traffic would have bypassed correction with a detector in place, thus creating a clear opportunity to significantly boost the computational efficiency for the entire system. In addition, a separate detector can also greatly benefit downstream search engine components. For example the detection probability score can be used to derive useful misspelling-confidence features for downstream rankers \cite{Agichtein06}\cite{Yin16} which rank search results to improve their ranking quality. From such perspectives, we clearly see the necessity and importance to build an independent misspelling detector. 

However, before deep learning and the recent advancement in contextual representations, for a noisy-channel based architecture, the efforts and complexity of building separate spell detector is equal to building the corrector since the features to build both are usually shared. What's even worse, our production experience shows that such a misspelling detector trained with similar features of the corrector ended up with overall lower correction quality and end user experience. In fact it's theoretically understandable since feature space duplication for mutually-coupled models would not be expected to improve the overall quality.

Now with the advancement on contextual representations such as BERTology, as well as availability of sufficient amount of quality training data by an in-house developed data algorithm breakthrough, building a state-of-the-art contextual misspelling detector becomes essential, critical and feasible for Maps serving instant search requests from global users.

With that being said, Maps search queries, with massive unique POI(Point-of-Interest) and AOI(Address-of-Interest) entities, significantly differ from common natural languages or public datasets such as Wikipedia. Could BERTology family still extend the success stories? 

In this study, we design 4 stages of misspelling detection models with comprehensive modeling choices and experiments. The models include basic LSTM models with varied embedding strategies, fine-tuned standard BERTology models such as BERT versus RoBERTa with pooling, cross-domain retrained BERT and fine-tuned models, as well as single-domain pre-trained BERT and fine-tuned slim models. 

Our results show advanced BERTology family model not necessarily better than standard BERT, and that cross-domain trained BERT which has incorporated contextual insights from giant natural language corpus does not necessarily outperform a domain-specific BERT, even a slim one. Our best model achieves F1 89.62\%,  surfacing a brand-new detector module for noisy-channel based system, also beating all newly implemented LSTM baseline models by 4-7\%.

\section{Related Work}
Similar to any classification problem, the key of building a misspelling detector for Maps search is to learn high quality representation of the input query. 

Earlier work include CNN \cite{Kalchbrenner14}, RNN \cite{Yogatama17} \cite{Ghosh17} \cite{Sakaguchi17} etc, which are regular deep learning architectures encoding a final representation with so-called hidden states. Such methods are usually empowered by embedding techniques, such as word2vec \cite{Mikolov13} and GloVe \cite{Pennington14}. 

Recently, contextual representation learning in sentence level with much-expanded exploration space based on attentions has made great progress. By learning common language representations by utilizing a large amount of unlabeled data, these models achieved state-of-the-art results, such as BERT \cite{Devlin18}, a multi-layer bidirectional Transformer \cite{Vaswani17} trained on large corpus with masked word prediction and next sentence prediction.
  
Building misspelling detector and serve it before corrector is essential, however detection compared with correction is much less discussed, and the importance of applying detection in real production associated with the most widely adopted noisy-channel approach is even yet disclosed. Some works discuss misspelling detection, correction and pattern analysis from language perspective, such as Arabic\cite{attia-etal-2012-improved}, Portuguese\cite{Priscila15}, Persian\cite{Yazdani20}, however the methods are either around noisy-channel paradigm or only language model or mainly the pattern analysis. We also see related discussions for domain-specific applications such as for clinical text\cite{Lai15}\cite{Wilbur06}\cite{Yazdani20}, and indeed search \cite{Martins04}\cite{Wilbur06}\cite{Popescu14} as well, however they either focus on patterns or improving a specific area under the noisy-channel architecture such as efficient dictionary or matching technique. Maps search is highly domain specific and significantly differs from natural languages, but building a misspelling detector for it is yet discussed, let alone leveraging BERTology, as well as exploring their domain adaptability. Could these technologies really move the needle? We need to tell with real-world data and comprehensive modeling experiments.

\section{Real-World Data}
The spell correction data in this paper are real-world data generated by an in-house developed data algorithm breakthrough. Table ~\ref{tab:datasamples}  shows a few data examples with other unrelated columns removed. We only briefly talk about the background and how the algorithm works since they are not the focus of the paper.

\begin{table}
  \caption{Data Samples}
  \label{tab:datasamples}
  \begin{tabular}{cccl}
    \toprule
    \textbf{Query} &\textbf{Correction} & \textbf{IsMisspelt} \\
    \midrule
    pondarosa auto & ponderosa auto & True \\
    6400 souu king & 6400 south king & True \\
    293 concord& 293 concord & False \\
    7789 southwest& 7789 southwest & False \\
    the portlander& the portlander & False \\
    liberty bowl& liberty bowl & False \\
    sni osle& sno isle & True \\
    \bottomrule
  \end{tabular}
\end{table}

\subsection{Background}

Traditional human based data generation is slow and expensive. In addition, 2-3\% noise can be often observed due to inevitable human mistakes. This greatly limits any speller related modeling explorations as well as scaling speller to all global locales for Maps search at production level.

\subsection{Algorithm}
The central mission for such an algorithm is to generate high quality pairs, Q(misspelt query) -> C(ground-truth correction), freely, accurately and at large scale, no dependency on human judgement. We successfully achieved such goal with an in-house designed novel distributed mining algorithm, which is a proved work in production. We briefly share how it works:

\subsubsection{Establish Recall}

\begin{itemize}
\item Temporal backtracking of user keystroke sequence: Locating most possible misspelling occurrences based on a set of signals given a temporal sequence, such as sequence curve, user engagement etc. 
\item Transfer mining: Capturing possible misspelling occurrences where the current search system already handles. Such misspellings should be incorporated since it’s an important part of the entire misspelling distribution.
\item The above two capture misspelt query candidates. For ground-truth corrections, it’s important to pay more attention to the user selected results instead of the user typed queries. Strong correlation and similarity between results and queries are underlying guaranteed by an existing auto-complete search system.
\end{itemize}

\subsubsection{Improve Precision}

For each pair of (Q, C), we scan forward and backward during the above recall phase to balance the morphological distance between Q and C not too far or too close based on adjustable heuristic rules. To do so, a few dynamic-programming methods can be employed, such as edit distance, longest common sub-sequence, etc.

\subsubsection{Ambiguity Resolution}
\begin{itemize}
\item Solving conflicts - Eliminate ambiguity when a single misspelt query points to multiple corrections in the mined data using majority voting algorithm based on the pair frequencies. 
\item Probability calibration - Boost the probability of a misspelt query to be corrected to a correction by also looking at the probability of the correction as a query without being corrected in the mined data.
\end{itemize}

\subsection{Data Overview}
As shown by the data examples, since we deal with classification, only Query and IsMisspelt are kept. The column of ‘Correction’ is only for analysis reference.

For this paper, we prepare 1 million examples for building the misspelling detector: 980,000 for training, 10,000 for dev and 10,000 for test. We use 20\% misspelling ratio which is the estimated upper bound as mentioned before to represent user query misspelling distribution. It's also worth noting that to maximize model's generalizability, we force no-overlapping between train, dev and test.

In addition, a 5 million dataset is also provided in preparation for the domain related BERTology model retraining. The misspelling ratio is increased to a balanced 50\% with the expectation to encourage the BERTology model to learn deeper misspelling insights.

\subsection{Preprocessing}
An existing implemented normalizer is applied on all the data. It suppresses all punctuations, removes junk characters such as emojis. It also collapses any multiple spaces into a single space between tokens.

\section{Models}
In this paper, we first build a set of baseline models leveraging the standard RNN/LSTM architecture, then explore various opportunities of leveraging BERTology family models for further improvements. We introduce 4 stages of models.

\subsection{Baseline Models}
Two new baseline models are implemented based on LSTM, with self-trainable word embeddings and external pre-trained embeddings such as GloVe. Performance with regard to embedding freeze is also evaluated. 

\subsubsection{LSTM with self-trainable embeddings} A classic LSTM model with 1 linear layer as output. Word level embeddings are self-learnable during learning. 
\subsubsection{LSTM with GloVe word embeddings} For this model, pre-trained GloVe embeddings will take the place of the self-trainable embeddings. 

\subsection{Fine-tuned BERT and RoBERTa}
Fine-tune pre-trained BERT and RoBERTa models (Hugging Face \footnote{\url{https://huggingface.co/}}) with the 1 million data examples with choices of hidden states, pooling strategies and other hyper-parameters.

\subsubsection{Regular fined-tuned BERT} We select only the last hidden state of the last output layer as the classification input. The motivation is to explore the effectiveness of the last hidden state. 

\subsubsection{Fine-tuned BERT with pooling average} We average the last hidden states of the last 4 output layers as the classification input. The motivation is to explore the average pooling of selected hidden states to compare with the default last hidden state from the last output layer.

\subsubsection{Regular fine-tuned RoBERTa} We explore if RoBERTa could bring more gains on top of BERT with the same data. 

\subsubsection{Regular fine-tuned RoBERTa with more hyper-parameter tuning} Our motivation is to explore a bit more on hyper-parameters to possibly improve model quality.

\subsection{Cross-domain Full BERT with Fine-tuning}
We retrain BERT from the publicly available pre-trained BERT using the 5 million augmentation dataset. We also evaluate the performance when freezing or unfreezing BERT for the fine-tuned models.

\subsection{Single-domain Slim BERT with Fine-tuning}
The cross-domain BERT allows BERT to learn certain domain-specific contexts to understand the domain task better. However, such BERT model is associated with pre-computed vocabulary including punctuation which can partially conflict with our data normalizer which does not allow punctuation. Also, the natural language corpus that trained the initial BERT also significantly differ from our domain data. Considering these factors, it’s important to train BERT from scratch with only the domain data, then fine-tune the target task on top. 

Specifically, we experiment a few slim BERT models in consideration of faster convergence and smaller data size (the same 5 million augmentation dataset as mentioned before) compared with the original giant natural language corpus, where we only keep half of the encoder layers of a full BERT. For the target fine-tuning, we train 3 models on top of the relatively longer-trained slim BERT, with further tuned hyper-parameters. 

\section{Experiments}

We evaluate all models with F1. 

\subsection{Baseline Models}

\subsubsection{LSTM with self-trainable embeddings}
The model has embedding dimension 50, hidden state dimension 50, and word vocabulary size 268,513, which is the max vocabulary size given the training data from the 1 million dataset.
The best model converges after 31 epochs with its best F1 0.820. We can see basic models such as LSTM actually works quite well.

\subsubsection{LSTM with GloVe word embeddings}
The first two models are trained with 6B Glove (Wikipedia and Gigaword) vectors, with the same embedding dimension 50, hidden state dimension 50, and the same vocabulary. The only difference is whether to freeze embeddings. The embedding-freeze model converges to the best model F1 0.782 after 28 epochs, and the embedding-non-freeze model reaches the best F1 0.840 after 41 epochs.

The second two models are trained with 42B Glove (Common Crawl) vectors, with embedding dimension 300, hidden state dimension 50 and the same vocabulary. Our motivation of using a different GloVe embedding set is Maps search data could be semantically closer to crawler data than Wikipedia data. We evaluate if the hypothesis is true. The results show that the embedding-freeze model reaches the best F1 0.784 after 25 epochs, and the embedding-non-freeze model reaches the best F1 0.858 after only 16 epochs.

\subsection{Fine-tuned BERT and RoBERTa}

In this experiment, we explore standard fine-tuning for BERT and RoBERTa. 

\subsubsection{Regular fined-tuned BERT} The model is a standard fine-tuned BERT using the last hidden state. The model achieves the best F1 0.8785 after epoch 3. The learning rate starts with 3e10-5 and dropout 0.3 for the linear layer. 

\subsubsection{Fine-tuned BERT with pooling average} The model is a standard fine-tuned BERT with average pooling of the hidden states of the last 4 layers, also with the same learning rate and dropout as the above model. The model achieves the same F1 0.8785 on the same epoch 3. Since pooling does not make a difference, we continue the strategy of only using the default last hidden state for the following models.

\subsubsection{Regular fine-tuned RoBERTa} This model is a fine-tuned RoBERTa instead of BERT, with max 3 epochs, batch size 32, learning rate 3e10-5, and dropout 0.3 on linear layer. The model F1 trends up over epochs but ends up with a final best F1 only 0.8484.  

\subsubsection{Regular fine-tuned RoBERTa with more hyper-parameter tuning} We explore further hyper-parameter tuning to conclude the model performance for RoBERTa. We use larger epochs 6 and batch size 128 , as well as different learning rates. We see the best F1 does increase to 0.8675, but still underperforming the model "Regular fined-tuned BERT".

\subsection{Cross-domain Full BERT with Fine-tuning}

\subsubsection{BERT non-freeze} The model reaches the best F1 0.8870 on epoch 2. The learning rate is 3e10-5 with dropout 0.3 on linear layer. However, the BERT pre-training takes significantly long with more than 10 hours.

\subsubsection{BERT freeze} The best model only reaches F1 0.79 on epoch 2 with the same learning rate and dropout as the non-freeze model. We can see that freezing BERT significantly worsens the target fine-tuned model.

\subsection{Single-domain Slim BERT with Fine-tuning}

\subsubsection{Slim BERT model with only domain data} We train a brand-new BERT with only domain data for maximum 3 epochs as a preliminary exploration considering training could take time. Then we train a fine-tuned model for max 4 epochs with learning rate 3e10-5 and dropout 0.3. The best model F1 achieves 0.8642 in this stage.

\subsubsection{Longer trained slim BERT} We increase the slim BERT training time to 10 epochs for longer learning. We keep the same hyper-parameters for the fine-tuned model and the best model F1 reaches 0.8891. We then freeze BERT for another round of fine-tuning, the F1 drops to 0.8300. Another evidence that BERT freeze harms the target model.
\begin{table}
  \caption{Best Model (Single-Domain Slim BERT with Fine-tuning)}
  \label{tab:bestmodel-single}
  \begin{tabular}{cccl}
    \toprule
    \textbf{Label} &\textbf{Precision} & \textbf{Recall} & \textbf{F1}\\
    \midrule
    1 & 0.8500 & 0.8167 & 0.8330 \\
    0 & 0.9549 & 0.9642 & 0.9595 \\
    \textbf{Macro Avg}  & 0.9024 & 0.8904 & \textbf{0.8962} \\
    \bottomrule
  \end{tabular}
\end{table}
\subsubsection{Longer trained slim BERT with additional hyper-parameter tuning} We experiment how other hyper-parameters would affect the fine-tuned model by increasing epochs to maximum 10 and reducing learning rate to 1e10-5, with the hope that small learning rate with longer iterations could further help. In this modeling iteration, we achieve the best model F1 0.8962 (Table ~\ref{tab:bestmodel-single}).

\section{Analysis}

\subsection{Metrics Analysis}

We analyze metrics for the above experiments and share our findings. All the representative models at each stage with best F1 scores are aggregated in a single Table ~\ref{tab:bestmodelsinarow}.

\begin{table}
  \caption{Best Models in a Row}
  \label{tab:bestmodelsinarow}
  \begin{tabular}{cl}
    \toprule
    \textbf{Model} &\textbf{Best F1}\\
    \midrule
    LSTM & 0.820\\
    LSTM+Glove Wiki+Gaga& 0.840\\
    LSTM+Glove Crawl& 0.858\\
    BERT Finetune& 0.8785\\
    Cross-domain full BERT & 0.8870\\
    Single-domain slim BERT & 0.8962\\
    \bottomrule
  \end{tabular}
\end{table}

\textbf{Baseline Models} show that LSTM with self-learned embeddings is able to establish a solid baseline with F1 0.82. Leveraging external pre-trained word embeddings such as GloVe further shows that retraining embeddings with the target task is usually a preferred choice, especially at production level where the domain data vocabulary is usually not small. In addition, the closer the semantic space between domain data and embedding source data, the more likely the model could further benefit from the external embeddings.

\textbf{Fine-tuned BERT and RoBERTa} show that 1. Variation of hidden states as representation fed into the target classification task does not make visible quality difference for the fine-tuning. 2. Fine-tuning RoBERTa is quite slower than BERT. 3. Longer epochs for fine-tuning RoBERTa does gradually improve F1 however it’s still not time friendly compared with BERT. 4. Even with additional hyper-parameter tuning, the best fine-tuned RoBERTa with F1 0.8675 still underperforms the best fine-tuned BERT with F1 0.8785. Therefore, for the following experiments, we discarded the option of RoBERTa. 

\textbf{Cross-domain Full BERT with Fine-tuning} shows the cross-domain non-freeze fine-tuned model further improves F1 by ~1\% (0.8785 -> 0.8870) compared with the standard non-freeze fine-tuned model. We see again freeze hurts model performance. Also, cross-domain BERT pre-training could take long. 

\textbf{Single-domain Slim BERT with Fine-tuning} shows that 1. Pre-training slim BERT with only domain data is significantly faster (less than a hour per epoch in our case) than pre-training cross-domain full BERT (hours per epoch in our case). 2. The new slim BERT trained from scratch with only domain data pushes the target model F1 by almost another ~1\% (0.8870 -> 0.8962) on top of the best cross-domain retrained full BERT. We strongly anticipate the single-domain pre-trained slim BERT could further improve the target fine-tuned model quality if given longer pre-training time. 3. Freeze hurts model quality again. 4. Hyper-parameters during fine-tuning such as longer training time with smaller learning rate could further benefit the target model. 

\subsection{Examples Analysis}

We analyze some success examples and failed examples for the best achieved model with F1 0.8962, as shown in Table ~\ref{tab:bestmodel-single}. Label True or 1 means misspelt, label False or 0 means non-misspelt. 

Given the dataset is unbalanced with 20\% misspelling ratio, the model does great on both the misspelt class (0.8330) and the non-misspelt class (0.9595). It's especially important for the model to perform strong on the weak class, which we have seen here the misspelt class, to be able to contribute significance to the overall strength of the model. 

\begin{table}
  \caption{Positive Examples}
  \label{tab:pos}
  \begin{tabular}{cccl}
    \toprule
    \textbf{Query} &\textbf{Label} & \textbf{Prediction} & \textbf{Correction}\\
    \midrule
    120 caiiage & True& True & 120 carriage\\
    weliwa springs& True & True& wekiva springs\\
    chilleens& False& False& chilleens\\
    puseyville& False& False& puseyville\\
    \bottomrule
  \end{tabular}
\end{table}

Let's first analyze a few examples where the model performs well in Table ~\ref{tab:pos}. We can see that Maps queries are highly domain specific. A word unit can be anything as long as it's part of a valid POI or AOI cross the globe. The model is able to detect common natural language misspelings, such as "caiiage" versus its expected correction “carriage”. It's also able to detect geographic terms that are challenging from natural language perspective, such as “weliwa”, “chilleens”, “pusevville” etc. These examples demonstrated the importance of pre-training domain-specific BERT model, and the effectiveness of such model's capability of representing contextual information within a specific domain.

\begin{table}
  \caption{Negative Examples}
  \label{tab:neg}
  \begin{tabular}{cccl}
    \toprule
    \textbf{Query} &\textbf{Label} & \textbf{Prediction} & \textbf{Correction}\\
    \midrule
    lily library & True& False & lilly library\\
    4430 park jazmin& False & True& 4430 park jazmin\\
    sunn fjord& False& True& sunn fjord\\
    \bottomrule
  \end{tabular}
\end{table}

Let’s also take a look at some failed cases in Table ~\ref{tab:neg}. We see something challenging here. The word “lily” is a correct word in common natural language, however when connecting to library under certain geographic intent, it may no longer be correct. The expected correction should be “lilly library” where “lilly” is a real and valid library name. For the remaining two examples, the model gets confused telling whether those queries are misspelt or not. Perhaps “jazmin”, “sunn”, and “fjord” would be where model gets confused. Those terms are all correct for the queries they belong to under Maps search domain, but they are unlike common words that are usually perceived from natural language perspective. For instance, people may have thought why not “jasmin”, “sun”, and “ford”, respectively. If the model tended to think the same way, it would lead to wrong predictions.

\section{Conclusion}

In this paper, we discussed importance and opportunity of building state-of-the-art domain-specific misspelling detector for Maps search. We implemented 4 stages of models by leveraging the most recent advancement in contextual representation learning and in-house data breakthrough in training data generation. These models include, basic embedding-powered LSTM models, the standard BERT and RoBERTa fine-tuned models, cross-domain full BERT fine-tuned models, and single-domain slim BERT fine-tuned models.

The experiments demonstrated improvements in the sequence of these models, with F1 0.82 increased to 0.8962. 

We summarize findings which could generalize to other similar domain-specific misspelling detection tasks, including but not limited to maps queries, clinical texts, and so on which reflect strong domain knowledge but significantly differ from natural language corpus the BERTology models have been trained on. 1. LSTM could create solid baseline and leveraging external embeddings can be further useful; 2. The last hidden state of BERT is empirically sufficient to represent full context; 3. Standard fine-tuned BERT can still outperform strong LSTM baseline; 4. Domain-specific fine-tuned BERT can outperform standard fine-tuned BERT for domain applications; 5. BERT is useful for resource-sparse target tasks given its capability to learn powerful contextual information during pre-training; 6. With more data, single-domain BERT trained from scratch should be able to further strengthen the target fine-tuning, even true for a slim version; 7. Data augmentation with balanced distribution could be a helpful strategy for unbalanced datasets; 8. Freezing BERT usually lowers the fine-tuning performance; 9. Hyper-parameters could help if following the right strategy. 

Here are some future plans to further improve the work: 1. Intervene target task learning with regression; 2. Incorporate Geohash\footnote{\url{https://en.wikipedia.org/wiki/Geohash}} to better address locality sensitive queries. For example, "genesys medical" and "genesis medical" are both valid depending on where the searches are performed. "1004 quary" and "1004 quarry" can be both valid as well.

\begin{acks}
We would like to thank Professor Chris Potts, Chair of Department of Linguistics from Stanford University, Powell Molleti from Stanford University, and Hari Bommaganti from Apple Inc, for the review and submission suggestions. We also would like to thank Alok Agarwal, Ankur Gupta, Max Muller III, and Patrice Gautier from Apple Inc for the support.
\end{acks}

\bibliographystyle{ACM-Reference-Format}
\bibliography{mybib}

\end{document}